\RequirePackage{fix-cm}
\documentclass[twocolumn]{svjour3}          
\smartqed  
\usepackage{graphicx}
\usepackage[%
  backend=bibtex      
 ,style=numeric-comp  
 ,sorting=nty        
 ,sortcites=true      
 ,block=none
 ,indexing=false
 ,citereset=none
 ,defernumbers=true
 ,isbn=false
 ,urldate=iso8601
 ,dateabbrev=true
 ,url=false
 ,doi=false            
 ,natbib=true         
 ,giveninits=true
]{biblatex}
\AtBeginBibliography{\small}

\usepackage{todonotes}
 \usepackage{wrapfig}

\usepackage[framemethod=tikz]{mdframed}
\definecolor{mycolor}{rgb}{0.122, 0.435, 0.698}

\newmdenv[innerlinewidth=0.5pt, roundcorner=4pt,linecolor=black,innerleftmargin=6pt, backgroundcolor=black!10,  
innerrightmargin=6pt,innertopmargin=6pt,innerbottommargin=6pt]{mybox}

\begin{document}

\title{From Chess and Atari to StarCraft and Beyond:\\ How  Game AI is Driving the World of AI}

\author{Sebastian Risi and Mike Preuss}

\institute{S. Risi \at
              IT University of Copenhagen and modl.ai \\
              Copenhagen, Denmark\\
              \email{sebr@itu.dk}           
           \and
           M. Preuss \at
             LIACS, Universiteit Leiden\\
             Leiden, Netherlands\\
              \email{m.preuss@liacs.leidenuniv.nl}
}

\maketitle

\begin{abstract}
This paper reviews the field of Game AI, which not only deals with creating agents that can play a certain game, but also with areas as diverse as creating game content automatically, game analytics, or player modelling. While Game AI was for a long time not very well recognized by the larger scientific community, it has established itself as a research area for developing and testing the most advanced forms of AI algorithms and articles covering advances in mastering video games such as StarCraft 2 and Quake III  appear in the most prestigious journals. Because of the growth of the field, a single review cannot cover it completely. Therefore, we put a focus on important recent developments, including that advances in Game AI are starting to be extended to areas outside of games, such as robotics or the synthesis of chemicals.           
In this article, we review the algorithms and methods that have paved the way for  these breakthroughs, report on the other important areas of Game AI research, and also point out exciting directions for the future of Game AI.
\end{abstract}

\section{Introduction}
For a long time, games research and especially research on Game AI was in a niche, largely unrecognized by the scientific community and the general public. Proponents of Game AI research wrote advertisement articles to justify the research field and substantiate the call for strengthening it (e.g. \cite{lucas2006evolutionary}). 
The main arguments have been these:

\begin{itemize}
    \item By tackling game problems as comparably cheap, simplified representatives of real world tasks, we can improve AI algorithms much easier than by modeling reality ourselves.
    \item Games resemble formalized (hugely simplified) models of reality and by solving problems on these we learn how to solve problems in reality.
\end{itemize}

Both arguments have at first nothing to do with games themselves but see them as a modeling\,/\,bench\-marking tools. In our view, they are more valid than ever. However, as in many other digital systems, there has also been and still is a strong intrinsic need for improvement because the performance of Game AI methods was in many cases too weak to be of practical use. This could be both in terms of playing strength, or simply because they failed to produce believable behavior~\cite{Livingstone2006}.
The latter would be necessary to hold up the suspension of disbelief, or, in other words, the illusion to willingly be immersed in a game world.

But what exactly is Game AI? Opinions on that have certainly changed in the last 10 to 15 years. For a long time, academic research and game industry were largely unconnected, such that neither researchers tackled AI-related problems game makers had nor the game makers discussed with researchers what these problems actually were.
Then, in research some voices emerged, calling for more attention for computer Game AI (partly as opposed to board game AI), including Nareyek~\cite{Nareyek2001, Nareyek2007}, Mateas~\cite{Mateas2003}, Buro~\cite{Buro2003}, and also Yannakakis~\cite{Yannakakis2012}.

Proponents of a change included Alex Champandard in his computational intelligence and games conference (CIG) 2010 tutorial \cite{YannakakisT11} and Youichiro Miyake in his GameOn Asia 2012 keynote\footnote{\url{http://igda.sakura.ne.jp/sblo_files/ai-igdajp/academic/YMiyake_GameOnAsia_2012_2_25.pdf}}.
At that time, a large part of Game AI research was devoted to board games as Chess and Go, with the aim to create the best possible AI players, or to game theoretic systems with the aim to better understand these.

Champandard and Miyake both argued that research shall try to tackle problems that are actually relevant also for the games industry. 
This led to a shift in the focus of Game AI research that was further intensified by a series of Dagstuhl meetings on Game AI that started in 2012\footnote{see \url{http://www.dagstuhl.de/12191}, \url{http://www.dagstuhl.de/15051}, \url{http://www.dagstuhl.de/17471}, \url{http://www.dagstuhl.de/19511}}.
The panoramic view \cite{YannakakisT15} explicitly lists 10 subfields and relates them to each other, most of which were not widely known as Game AI at that time, and even less so in the game industry. Most prominently, areas with a focus on using AI for design and production of games emerged, such as procedural content generation (PCG), computational narrative (nowadays also known as interactive storytelling), and AI-assisted game design. Next to these, we find search and planning, non-player character (NPC) behavior learning, AI in commercial games, general Game AI, believable agents, and games as AI benchmarks.
A third important branch that came up at that time (and resembles the 10th subfield) considers modeling players and understanding what happens in a running game (game analysis).

The 2018 book on AI and Games \cite{gameAIbook} shows the pre-game (design\,/\,production) during game (game playing) and after-game (player modeling\,/\,game analysis)\footnote{We are aware that this division is a bit simplistic, of course players can be also modeled online or for supporting the design phase. Please consider this a rough guideline only.} 
uses of AI together with the most important algorithms behind it and gives a good overview of the whole field. Due to space restrictions, we cannot go into details on developments in each sub-area of Game AI in this work but rather provide an overview over the ones considered most important, including highlighting some amazing recent achievements that for a long time have not been deemed possible. These are mainly in the game playing field but also draw from generative approaches such as PCG in order to make them more robust.

Most of the popular known big recent successes are connected to big AI-heavy IT companies entering the field such as DeepMind (Google), Facebook AI and OpenAI. Equipped with rich computational and human resources, these new players have especially profited from Deep (Reinforcement) Learning to tackle problems that were previously seen as important milestones for AI, successfully tackling difficult problems of human decision making, such as Go, Dota2, and StarCraft.

It is, however, a fairly open question how we can utilize these successes for solving other problems in Game AI and beyond. As it appears to be possible but utterly difficult to transfer whole algorithmic solutions, e.g., for a complex game as StarCraft, to a completely different domain, we may rather see innovative recombinations of algorithms from the recently enriched portfolio in order to craft solutions for new problems.  

In the next sections, we start with enlisting some important terms that will be repeatedly used (Sect.\ref{sec:terms}) before tackling state\,/\,action based learning in Sect.~\ref{sec:states}.
We then report on pixel-based learning in Sect.~\ref{sec:pixels}. At this point, PCG comes in as a flexible testbed generator (Sect.~\ref{sec:pcg}). However, it is also a viable aim on its own to be able to generate content.
Very recently, different sources of game information, such as pixel and state information, are given as input to these game-playing agents, providing better methods for rather complex games (Sect.~\ref{sec:pixelstates}). While many approaches are tuned to one game, others explicitly strive for more generality (Sect.~\ref{sec:agi}). Next to game playing and generating content, we also shortly discuss AI in other roles (Sect.~\ref{sec:other}). We conclude the article with a short overview of the most important publication venues and test environments in Sect.~\ref{sec:venues} and some reasoning about the 
expected future developments in Game AI in Sect.~\ref{sec:future}.
\vspace*{2ex}

\section{Algorithmic approaches and game genres}
\label{sec:terms}

We provide an overview of the predominant paradigms\-\,/\,algorithm types and game genres, focusing mostly on game playing and more recent literature. These algorithms are used in many other contexts of AI and application areas of course, but some of their most popular successes have been achieved in the Game AI field.
\medskip

\begin{mybox}
\textbf{Reinforcement Learning (RL).} In reinforcement learning an agent learns to perform a task through interactions with its environment and through rewards. This is in contrast to supervised learning, in which the agent is directly told the correct action in different states. One of the main challenges in RL is to find a balance between exploitation (i.e.\ seeking out states that are known to give a high reward) vs. exploration (i.e.\ trying out something new that might lead to higher rewards in the long run). 
\end{mybox}

\pagebreak[3]

\begin{mybox}
\textbf{Deep Learning (DL).}  Deep learning is a broad term and comes in a variety of different shapes and sizes. The main distinguishing feature of deep learning is the idea to learn progressively higher-level features through multiple layers of non-linear processing. The most prevalent deep learning methods are based on deep neural networks, which are artificial neural networks with multiple different layers (in new neural network models these can be more than 100 layers). Recent advances in computing power, such as more and more efficient GPUs (which were first developed for fast rendering of 3D games), more data, and various training improvements have allowed deep learning methods to surpass the previous state-of-the-art in many domains such as image recognition, speech recognition or drug discovery. LeCun et al.~\cite{lecun2015deep} provide a good review paper on this fast-growing research area. 
\end{mybox}

\begin{mybox}
\textbf{Deep Reinforcement Learning.} Deep Reinforcement Learning combines reinforcement learning with deep neural networks to create efficient algorithms that can  learn directly from high-dimensional sensory streams. Deep RL has been the workhorse behind many of the recent advances in Game AI, such as beating professional players in StarCraft and Dota2. 
\cite{arulkumaran2017brief} provides a good overview over deep RL. 
\end{mybox}

\begin{mybox}
\textbf{Monte Carlo Tree Search (MCTS).}
Monte Carlo Tree Search is a fairly recent \cite{Coulom06} randomized tree search algorithm. States of the mapped system are nodes in the tree, and possible actions are edges that lead to new states.
In contrast to older methods such as alpha-beta pruning, it does not attempt to look at the full tree but uses controlled exploration and exploitation of already obtained knowledge (successful branches are preferred) and often fully randomized playouts, meaning that a game is played until it ends by applying randomized actions. If that takes too long, state value heuristics can be used alternatively. Loss/win information is propagated upwards up to the tree root such that estimations of the win ratio at every node get available for directing the search. MCTS can thus be applied to much larger trees, but provides no guarantees concerning obtaining optimal solutions.  \cite{Browne2012} is a popular introductory survey.
\end{mybox}

\pagebreak[3]
\vspace*{2ex}

\begin{mybox}
\textbf{Evolutionary Algorithms (EA).} 
Also known as bio-inspired optimization algorithms, Evolutionary Algorithms take inspiration from natural evolution for solving black-box optimization problems. They are thus applied when classical optimization methods fail or cannot be employed because no gradient or not even numeric objective value information (but ranking of solutions) is available. 
A key idea of EAs is parallel search by means of populations of candidate solutions, which are concurrently improved, making it a global optimization method. EAs are especially well suited for multi-objective optimization, and the well-known GA, NSGA-II, CMA-ES algorithms are all EAs, see also the introduction/survey book \cite{Eiben2015}.
\end{mybox}

\begin{figure}[t]
    \centering
    \includegraphics[width=0.76\columnwidth]{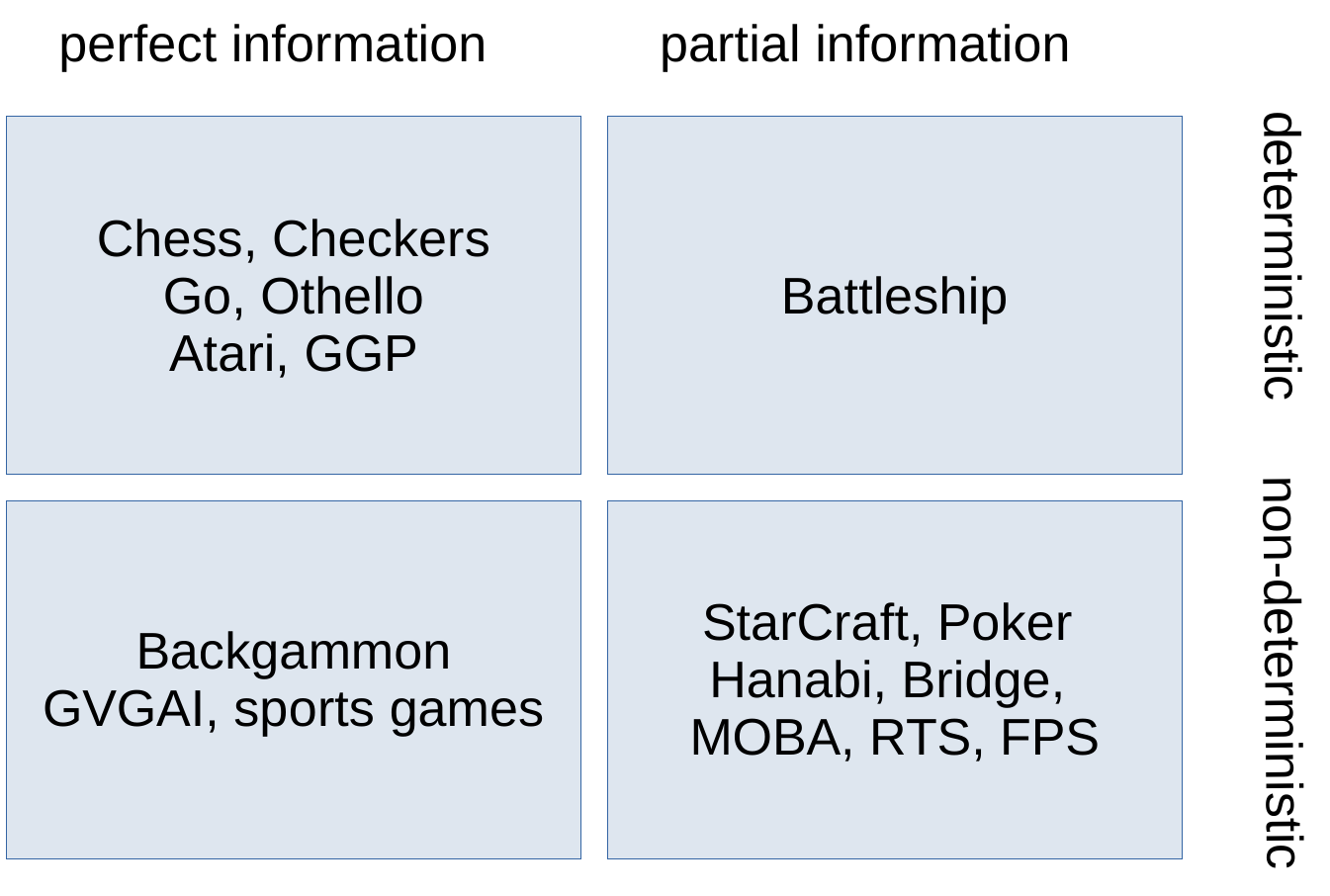}
    \caption{Available information and determinism as separating properties for different games treated in Game AI.}
    \label{fig:determinism}
\end{figure}

\begin{figure}[t]
    \centering
    \includegraphics[width=0.98\columnwidth]{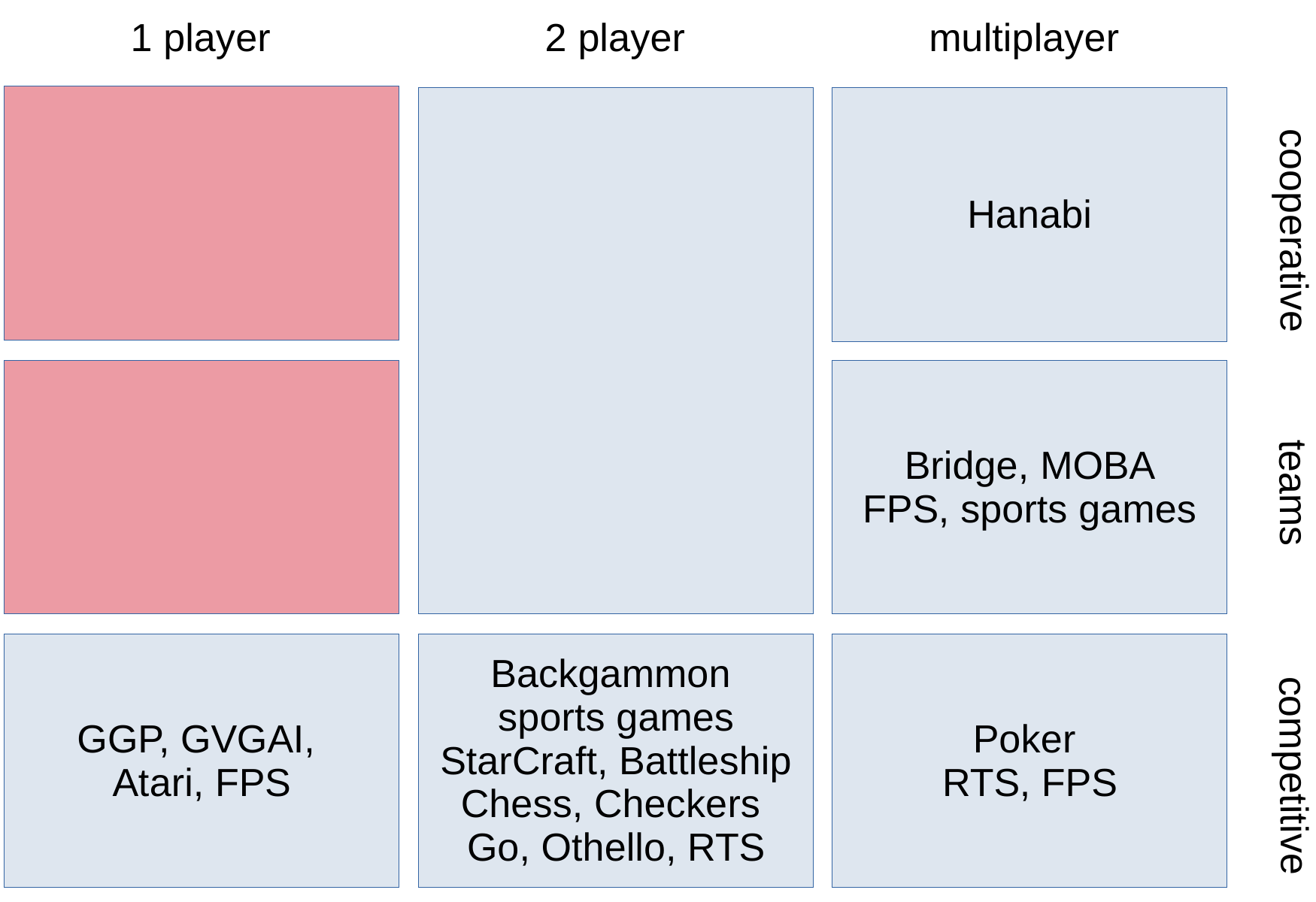}
    \caption{Player numbers and style from cooperative to competitive for different games or groups of games treated in Game AI. Note that for several games, multiple variants are possible, but we use only the most predominant ones. }
    \label{fig:cooperative}
\end{figure}

Which are the most important games to serve as testbeds in Game AI? The research-oriented frameworks general game playing (GGP), general video Game AI (GVGAI) and the Atari learning environment (ALE) play an important role but are somewhat far from modern video games.
This also holds true for the traditional AI challenge board games Chess and Go and card games as Poker or Hanabi.
In video games,  the predominant genres are 
real-time strategy (RTS) games such as StarCraft, Multiplayer online battle arena (MOBA) games such as Dota2, and first person shooter (FPS) games such as Doom. Sports games currently get more important \cite{liu2019} as they often represent a competitive team situation that is seen as similar to many real-world human/AI collaborative problems. In a similar way, 
cooperative (capture-the-flag) variants of FPS games \cite{jaderberg2019human} are used. Figures~\ref{fig:determinism} and \ref{fig:cooperative} provide an overview of the different properties of the games used as AI testbeds.

\section{Learning to play from states and actions}
\label{sec:states}

Games have for a long time served as invaluable testbeds for research in artificial intelligence (AI). In the past, particularly board games such as Checkers and Chess have been tackled, later on turning to Go when Checkers had been solved \cite{Schaeffer2007}
and with DeepBlue \cite{campbell2002deep} an artificial intelligence had defeated the world champion in Chess consistently. All these games and many more, up to Go, have one thing in common: they can be expressed well by states and actions, where the number of actions is usually a not-too-large number of often around 100 or less reasonable moves from any possible position. For quite some time, board games have been tackled with alpha-beta pruning (Turing Award Winners Newell and Simon explain in \cite{newellturing} how this idea came up several times almost at once) and very sophisticated and extremely specialized heuristics before Coulom invented Monte Carlo Tree Search (MCTS)
\cite{Coulom06} in 2006. MCTS gives up optimality (full exploration) in exchange for speed and is therefore now dominating AI solutions for larger board games such as Go with about $10^{170}$ possible states (board positions).
MCTS-based Go algorithms had greatly improved the state-of-the-art up to the level of professional players by incorporating sophisticated heuristics as Rapid Action Value Estimation (RAVE) \cite{Gelly2011}. In the following, 
MCTS based approaches were shown to cope well also with real-time conditions as in the PacMan game \cite{Pepels2014} and also hidden information games \cite{Powley2014}.

However, only the combination of MCTS with DL led to a world-class professional human-level Go AI player named AlphaGo \cite{Silver2016}.
At this stage, human experience (recorded grandmaster games) had been used for "seeding" the learning process that was then accelerated by self-play. By playing against itself, the AlphaGo algorithm was able 
to steadily improve its value (how good is the current state?) and policy (what is the best action to play?) artificial neural networks.
The next step, AlphaGo Zero \cite{silver2017mastering}  removed all human data, relying on self-play alone, and learned to play Go better than the original AlphaGo approach but from scratch. 
This approach has been further developed to AlphaZero \cite{Silver2018}
and shown to be able to learn to play different games, next to Go also Chess and Shogi (Japanese Chess).
In-depth coverage of  most of these developments is also provided in \cite{Plaat2020}\footnote{\url{https://learningtoplay.net/}}.

From the last paragraphs, it may appear as if learning via self-play is limited to two-player perfect information games only. However, also multi-player partial information games such as Poker~\cite{Brown2019} and even cooperative multi-player games such as Hanabi~\cite{Lerer2019} have recently been tackled and AI players now exist that can play these games at the level of the best human players.
Thus, is self-play the ultimate AI solution for all games? Seemingly not, as
\cite{vinyals2019} suggests (see Sect.~\ref{sec:pixelstates}). However, this may be a question of the number of actions and states in a game and remains to be seen. Nevertheless, board games and card games are obviously good candidates for such AI approaches.

\begin{figure*}
    \centering
    \includegraphics[width=2.0\columnwidth]{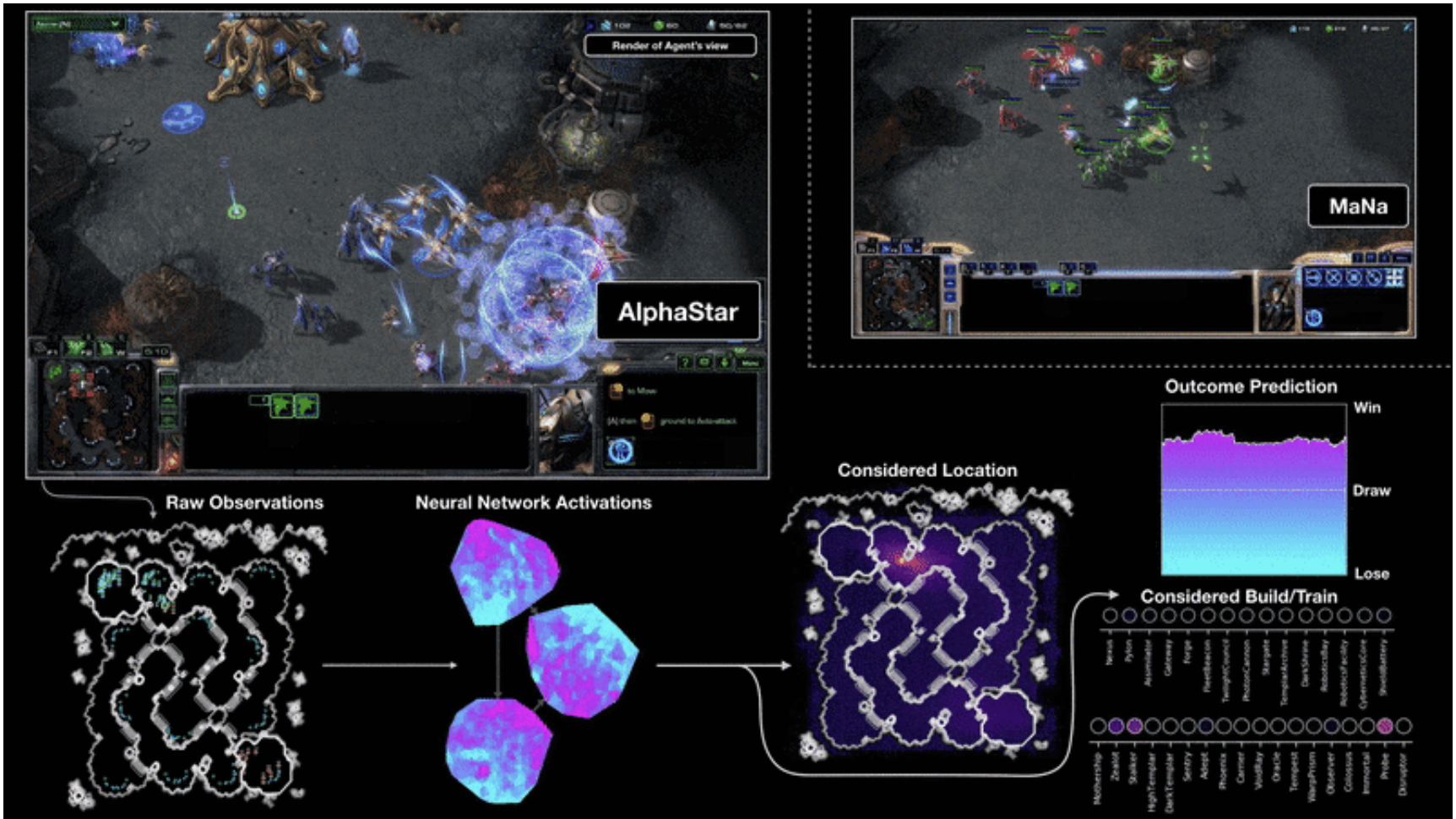}
    \caption{A visualisation of the AlphaStar agent playing against the human player MaNa, from \cite{alphastarblog}. Shown is the raw observation that the neural network gets as input (bottom left), together with the internal neural network activations. On the lower right side are shown actions considered by the agent together with a prediction of the outcome of the game.  }
    \label{fig:alphastar}
\end{figure*}

\section{Learning to play from pixels} 
\label{sec:pixels}

For a long time, learning directly from high-dimensional input data such as the pixels of a video game was an unsolved challenge.  Earlier neural network-based approaches for playing games such as Pac-Man relied on careful  engineered features such as the distance to the nearest ghost or pill, which are given as input to the neural network \cite{risi2015neuroevolution}. 

While some earlier game-playing approaches, especially from the evolutionary computation community, showed initial success in learning directly from pixels \cite{parker2012neurovisual,gallagher2007evolving,togelius2009super,hausknecht2014neuroevolution}, it was not until DeepMind's seminal paper on learning to play Atari video games from pixels \cite{Mnih2013,mnih2015human} that these approaches started to compete and at times outperform human players. Serving as a common benchmark, many novel AI algorithms have been developed and compared on Atari video games first \cite{justesen19review} before being applied to other domains such as robotics \cite{akkaya2019solving}. 
A computationally cheap and thus interesting end-to-end pixel-based learning environment is VizDoom~\cite{kempka2016vizdoom}, a competition setting that relies on a rather old game that is run in very small screen resolutions. Low resolution pixel inputs are also employed in the obstacle tower challenge (OTC) \cite{Juliani2019}.

DeepMind's paper ushered in the area of \emph{Deep Reinforcement Learning}, combining reinforcement learning with a rich neural network-based representation (see infobox for more details). Deep RL has since established itself as the prevailing paradigm is to learn directly from high-dimensional input such as images, videos, or sounds without the need for human-design features or preprocessing. More recently, approaches based on evolutionary algorithms have shown to also be competitive with approaches based on gradient descent-based methods \cite{such2017deep}.

However, some of the Atari games, namely Montezuma's Revenge, Pitfall, and others proved to be too difficult to solve with standard deep RL approaches \cite{mnih2015human} because of sparse and/or late rewards. These hard-exploration games can be handled successfully by evolutionary  algorithms that explicitly favor exploration such as Go-Explore~\cite{ecoffet2019go}.

A recent trend in deep RL is to allow agents to learn a general model of how their environment behaves and use that model to explicitly plan ahead. For games, one of the first approaches was the World Model introduced by  
\cite{ha2018world}, in which an agent learns to solve a challenging 2D car racing game and a 3D VizDoom environment from pixels alone. In this approach, the agent first learns by collecting observations from the environment, and then training a forward model that takes the current state of the environment and action and tries to predict the next state. Interestingly, this approach also allowed an agent to get better by training inside a hallucinated environment created through a trained world model. 

Instead of first training a policy on random rollouts, follow-up work showed that end-to-end learning through  reinforcement learning \cite{hafner2018learning} and evolution \cite{risi2019gecco,risi2019improving}
is also possible. We will discuss MuZero as another example of planning in latent space in Section~\ref{sec:pixelstates}. 

\section{Procedural content generation} 
\label{sec:pcg}

In addition to playing games, another active area of AI research is procedural content generation (PCG) \cite{PCGbook,risi2019procedural}. PCG refers to the algorithmic creation of game content such as levels, textures, quests, characters, or even the rules of the game itself.  

One of the appeals of employing PCG in games is that it can increase their replayability by offering the player a new experience every time they play. For example, games such as No Man's Sky (Hello Games, 2016) or Spelunky (Mossmouth,
LLC, 2013) famously featured PCG as part of their core gameplay, allowing players to explore an almost unlimited variety of planets or caves. One of the most important early benefits of PCG methods was that it allowed the creation of larger game worlds than what would normally fit on a computer's hard disk at the time.  One of the first games using PCG-based methods was  \emph{Elite} (Brabensoft, 1984), a space trading video game featuring thousands of planets. The whole starsystem with each visited planet and space stations could be recreated from a given random seed.

While the origin of PCG is rooted in creating a more engaging experience for players~\cite{yannakakis2011experience}, more recently PCG-based approaches have also found important other use cases. With the realisation that methods such as deep reinforcement learning can surpass humans in many games, also came the realisation that these methods overfit to the exact environment they are trained on \cite{justesen2018illuminating,zhang2018study}. For example, an agent trained to reach the level of a human expert in a game such as Breakout, will fail completely when tested on a Breakout version where the game pedal has a slightly different size or is at a slightly different position. Recent research showed that by training agents on many procedurally generated levels allows them to become significantly more general \cite{justesen2018illuminating}. In an impressive extension of this idea, DeepMind trained agents on a large number of randomly created levels to reach human-level performance in the Quake III Capture the Flag game \cite{jaderberg2019human}. This trend to make AI approaches more general by training them on endless variations of environments was continued in the hide-and-seek work by OpenAI~\cite{Baker2019} and also in the obstacle tower challenge (OTC) \cite{Juliani2019} and will certainly also be employed in many future approaches.

Meanwhile, PCG has been applied to many different types of game components or facets (e.g.\ visuals, sound), but most often to only one of these at once. One of the open research questions in this context is how generators for different facets can be combined~\cite{Liapis2019}.

Similar to some of the other techniques described in this article, PCG has also more recently found to be applicable to areas outside of games \cite{risi2019procedural}. For example, training a humanoid robot hand to manipulate a Rubik's cube in a simulator on many variants of the same problem (e.g.\ varying parameters such as the size, mass, and texture of the cube) has allowed a policy trained in a simulator to sometimes work on a physical robot hand in the real world. For a review of how PCG has increased generality in machine learning we refer the interested reader to this survery \cite{risi2019procedural} and for a more in-depth review of PCG in general to the book by Shaker et al.~\cite{PCGbook}.

\section{Merging state and pixel information}
\label{sec:pixelstates}

Whereas the AI in AlphaGo and its predecessors for playing board games dealt with board positions and possible moves, deep RL and recent evolutionary approaches for optimising deep neural networks (a research field now referred to as deep neuroevolution \cite{stanley2019designing}), learn to play Atari games directly from pixel information. On the one hand, these approaches have some conceptual simplicity, but on the other hand, it is intuitively clear that adding more information -- if available -- may be of advantage.
More recently, these two ways of obtaining game information were joined in different ways. 

The hide-and-seek approach \cite{Baker2019} depends on visual and state information of the agents but also heavily on the use of co-evolutionary effects in a multi-agent environment that very much reminds of EA techniques.

In AlphaStar (Fig.~\ref{fig:alphastar}) that was designed to play StarCraft at human professional level, both state information (location and status of units and buildings) as well as pixel information (minimap) is fed into the algorithm.
Interestingly, self-play is used heavily, but is not sufficient to generate human professional competitive players because the strategy space is huge and human opponents may come up with very different ways to play the game that must all be handled.
Therefore, as in AlphaGo, human game data is used to seed the algorithm. Furthermore, also co-evolutionary effects in a 3 tier league of different types of agents are driving the learning process. It shall be noted that the success of AlphaStar was hard to imagine only some years ago because RTS games were considered the hardest possible testbeds for AI algorithms in games \cite{Ontanon2013}. These successes are, however, not without controversy and people argue if the comparisons of AIs playing against humans are fair \cite{justesen2019we,canaan2019leveling}.

MuZero \cite{schrittwieser2019mastering} is able to learn playing Atari games (pixel input) as well as Chess and Go (state input) by generating virtual states according to reward/position value similarity. These are managed in a tree-like fashion as in MCTS but costly rollouts are avoided. The elegance of this approach lies in the ability to use different types of input and the construction of an internal representation that is oriented only at values and not at exact game states.

\section{Towards more general AI}
\label{sec:agi}
While AI algorithms have become exceedingly good at playing specific games \cite{justesen19review}, it is still an unsolved challenge how to make an AI algorithm that can learn to quickly play any game it is given, or how to transfer skills learned in one game to another. This challenge, also known as General Video Game Playing \cite{genesereth2005general}, has resulted in the development of the 
 General Video Game AI framework (GVGAI), a flexible framework designed to facilitate the development of general AI through video game playing \cite{perez2016general}. 

With increasingly complicated worlds and graphics, video games might be the ideal environment to learn more general intelligence. Another benefit of games is that they often share similar controllers and goals. To spur developments in this area, the GVGAI framework now also includes a Learning Track, in which the goal of the agent is to learn a new game quickly without being trained on it beforehand. The hope is that methods that can quickly learn any game they are given, will also ultimately be able to quickly learn other tasks such a robot manipulation in the real world. 

Whereas most successful approaches for GVGAI games employ MCTS, it shall be noted that there are also other competitive approaches as the rolling horizon evolutionary algorithm (RHEA) \cite{PerezSLR13} that evolve partial action sequences as a whole through an evolutionary optimization process. Furthermore, DL variants start to get used here as well~\cite{torrado2018}.

\section{AI for player modelling and other roles}
\label{sec:other}

In this section, we briefly mention a few other use cases for current AI methods.  In addition to learning to play or generating games and game content, another important aspect of Game AI -- and potentially currently the main use case in the game industry -- is game analytics. Game analytics has changed the game landscape dramatically over the last ten years. The main idea in game analytics is to collect data about the players while they play the game and then update the game on the fly. For example, the difficulty of levels can be adjusted or the user interface can be streamlined. At what point players stopped playing the game can be an important indication of what to change to reduce the game's churn\footnote{In the game context, churn means that a player who has played a game for some time completely stops playing it. This is usually very hard to predict but essential to know especially for online game companies.} rate \cite{runge2014churn,hadiji2014predicting,Kummer2018}. 
We refer the interested reader to the book on game analytics by El-Nasr et al.~\cite{el2016game}. 

Another important application area of Game AI is player modelling. As the name suggests, player modelling aims to model the experience or behavior of the player \cite{bakkes2012player,yannakakis2013player}. One of the main motivations for learning to model players is that a good player model can allow the game to be tailored even more to the individual player. A variety of different approaches to model players exist, such as supervised learning (e.g.\ training a neural network in a supervised way on recorded plays of human players to behave the same way), to unsupervised approaches such as clustering that aim to group similar players together \cite{drachen2009player}. Based on which cluster a new player belongs to, different content or other game adaptations can be performed. Combining PCG (Sect.~\ref{sec:pcg}) with player modelling, an approach called Experience-Driven Procedural Content Generation \cite{yannakakis2011experience}, allows  these algorithms to automatically generate unique content that induces a desired experience for a player. For example, \cite{pedersen2010modeling} trained a model on players of Super Mario, which could then be used to automatically generate new  Mario levels that maximise the modelled fun value for a particular player. Exciting recent work can even predict a player's affect in certain situation from pixels alone \cite{makantasis2019pixels}. 

There is also a large body of research on human-like non-player characters (NPC) \cite{hingston2012}, and some years ago, this research area was at the core of the field, but with the upcoming interest in human/AI collaboration it is likely to thrive again in the next years.

Other roles for Game AI include playtesting and balancing which both belong to game production and mostly happen before games are published. Testing for bugs or exploits in a game is an interesting application area of huge economic potential and some encouraging results exist \cite{denzinger2005dealing}. With the rise of machine learning methods that can play games at a human or beyond human level and methods that can solve hard-exploration games such as Montezuma's Revenge \cite{ecoffet2019go}, this area should see a large increase of interest from the game industry in the coming years.  Mixed-initiative tools that allow humans to create game content together with a computational creator often include an element of automated balancing, such as balancing the resources on a map in a strategy game \cite{liapis2013sentient}. 
Game balancing is a wide and currently under-researched area that may be understood as a multi-instance parameter tuning problem. One of the difficulties here is that many computer games do not allow headless accelerated games and APIs for controling these. Some automated approaches exist for single games~\cite{PreussPVP18} but they usually cannot cope with the full game and approaches for more generally solving this problem are not well established yet \cite{Volz2019}.
Dynamic re-balancing during game runtime is usually called dynamic difficulty adaptation (DDA) \cite{spronck2006adaptive}. 

\begin{figure*}
    \centering
    \includegraphics[width=1.5\columnwidth]{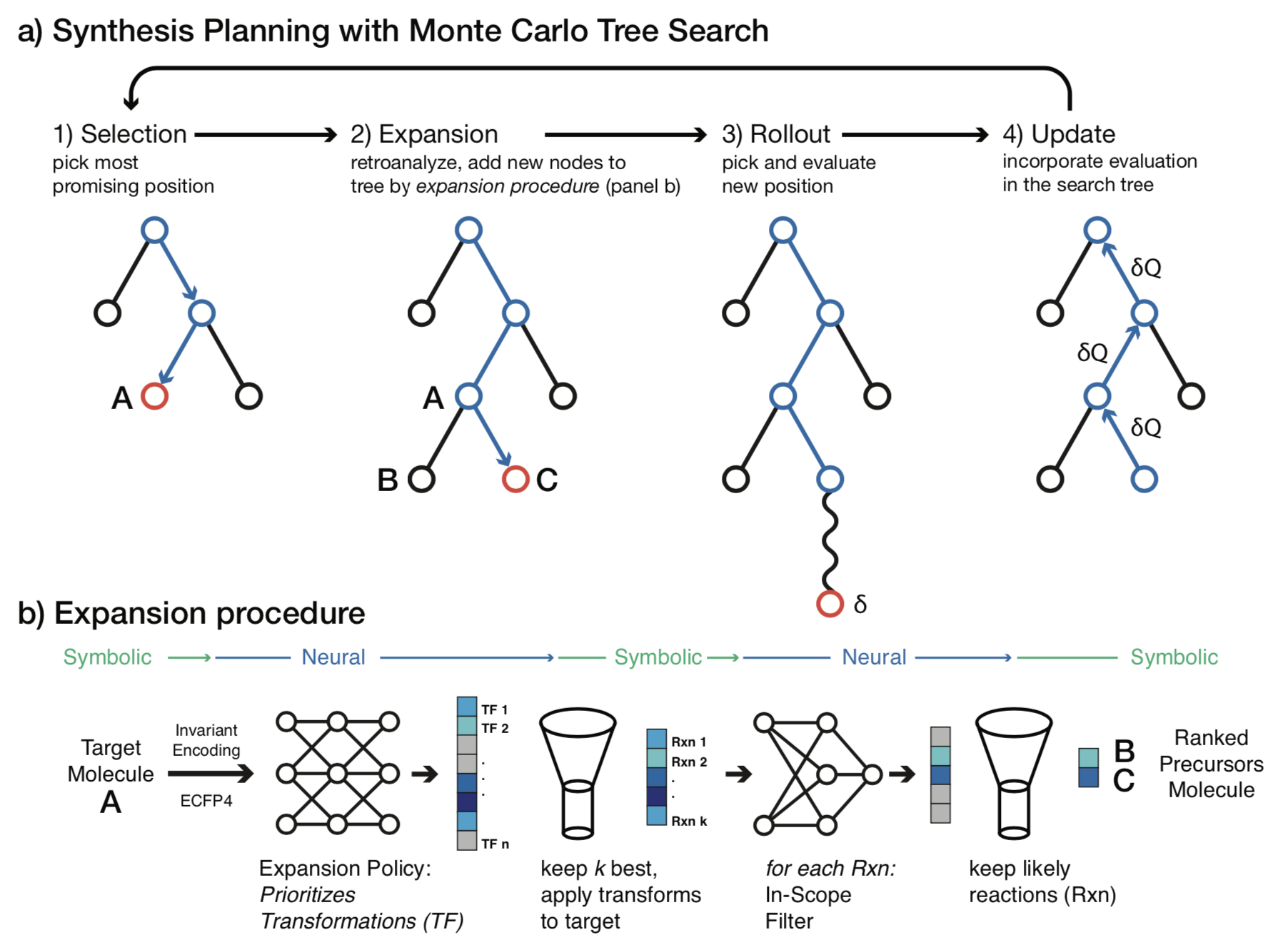}
    \caption{Chemical retrosynthesis on basis of the AlphaGo approach; figure from \cite{segler2018planning}. The upper subfigure shows the usual MCTS steps, and the lower subfigure links these steps to the chemical problem. Actions are now chemical reactions, states are the derived chemical compounds. Instead of preferred moves in a game, the employed neural networks learn reaction preferences. In contrast to AlphaGo, possible moves are not simply provided but have to be learned from data, an approach termed "world program"\cite{segler2019world}.
    \label{fig:chem} }
\end{figure*}

\section{Journals, conferences, and competitions}
\label{sec:venues}

The research area of Game AI is centered in computer science, but influenced by other disciplines as i.e. psychology, especially when it comes to handling humans and their emotions \cite{yannakakis2014emotion, yannakakis2018ordinal}. 
Furthermore, (computational) art and creativity (for PCG), game studies (formal models of play) and game design are important neighboring disciplines.

In computer science, Game AI is not only limited to machine learning and traditional branches of AI but also has  
links to information systems, optimization, computer vision, robotics, simulation, etc.
Some of the core conferences for Game AI are:
\begin{itemize}
    \item Foundations of Digital Games (FDG)
    \item IEEE Conference on Games (CoG), until 2018 the Conference on Computational Intelligence and Games (CIG)
    \item Artificial Intelligence for Interactive Digital Entertainment (AIIDE)
\end{itemize}

Also, many computer science conferences have tracks or co-located smaller conferences on Game AI, as e.g. GECCO and IJCAI. 
The more important journals in the field are the  
IEEE Transactions on Games ToG (formerly TCIAIG) and the 
IEEE Transactions on Affective Computing. The most active institutes in the area can be taken from a list (incomplete, focused only on the most relevant venues) compiled by Mark Nelson.\footnote{\url{http://www.kmjn.org/game-rankings}}

A large part of the progress of the last years is due to the free availability of competition environments as:
StarCraft, GVGAI, Angry Birds, Hearthstone, Hanabi, MicroRTS, Fighting Game, Geometry Friends and more, and also the 
more general frameworks as:
ALE, GGP, OpenSpiel, OpenAIGym, SC2LE, MuJoCo, DeepRTS.

\section{The future of Game AI}
\label{sec:future}

More advanced AI techniques are slowly finding their way into the game industry and this will likely increase significantly over the coming years. Additionally, companies are more and more collaborating with research institutions, to bring the latest innovations out to the industry. For example, Massive Entertainment and the University of Malta collaborated to predict the motivations of players in the popular  game Tom Clancy’s The Division \cite{melhart2019your}. Other companies, such as King, are investing heavily in  deep learning methods to automatically learn models of players that can then be used for playtesting new levels quickly \cite{gudmundsson2018human}.

Procedural content generation is already employed for many mainstream games such as Spelunky (Mossmouth,
LLC, 2013)  and No Man's Sky (Hello Games, 2016) and we will likely see completely new types of games in the future that would be impossible to realise without sophisticated AI techniques. The recent AI Dungeon 2 game (www.aidungeon.io) points to what type of direction these games might take. In this text adventure game players can interact with  Open AI's GPT-2 language model, which was trained on 40 gigabytes from text scraped from the internet. The game responds to almost anything the player types in a sensible way, although the generated stories  also often lose coherence after a while. This observation points to an important challenge: For more advanced AI techniques to be more broadly employable in the game industry, approaches are needed that are more controllable and potentially interpretable by designers \cite{zhu2018explainable}. 

We predict that in the near future, generative modelling techniques from machine learning, such as Generative and Adversarial Networks (GANs) \cite{goodfellow2014generative}, will allow users to personalise their avatars to an unprecedented level or allow the creation of an unlimited variety of realistic textures and assets in games. This idea of \emph{Procedural Content Generation via Machine Learning} (PCGML) \cite{summerville2018procedural}, is a new emerging research area that has already led to promising results in generating levels for games such as Doom \cite{giacomello2018doom} or Super Mario \cite{volz2018evolving}.

From the current perspective, we would expect that future research (next to playing better on more games) in Game AI will focus on these areas:
\begin{itemize}
    \item AI/human collaboration and AI/AI agent collaboration is getting more important, this may be subsumed under the term \emph{team AI}. Recent attempts in this direction include e.g.: Open AI five~\cite{raiman2019long}, Hanabi~\cite{Bard2019}, capture the flag~\cite{jaderberg2019human}
    \item More natural language processing enables better interfaces and at some point free-form direct communication with game characters. Already existing commercial voice-driven assistance systems as the Google Assistant or Alexa show that this is possible.
    \item The previous points and the progress in player modeling and game analysis will lead to more human-like behaving AI, this will in turn enable better playtesting that can be partly automated. 
    \item PCG will be applied more in the game industry and other applications. For example, it is used heavily in Microsoft's new flight simulator version that is now (January 2020) in alpha test mode.
    This will also trigger more research in this area.
\end{itemize}

Nevertheless, as in other areas of artificial intelligence, Game AI will have to cope with some issues that mostly stem from two newer developments: theory-light but very successful deep learning methods, and highly parallel computation. The first entails that we have very little control over the performance of deep learning methods, it is hard to predict what works well with which parameters, and the second one means that many experiments can hardly ever be replicated due to hardware limitations. E.g., Open AI Five has been trained on 256 GPUs and 128,000 CPUs~\cite{OpenAI_dota} for a long time. More generally, large parts of the deep learning driven AI are currently presumed to run into a 
reproducibility crisis\footnote{\url{https://www.wired.com/story/artificial-intelligence-confronts-reproducibility-crisis/}}. Some of that can be cured by better experimental methodology and statistics as also worked well in Evolutioanry Computation some time ago~\cite{bartz2010experimental}. First attempts in Game AI also try to approach this problem by defining guidelines for experimentation, e.g.\ for the ALE~\cite{Machado2017}, but replicating experiments that take weeks is an issue that will probably not easily be solved.

It is definitively desired to apply the algorithms that successfully deal with complex games also to other application areas. Unfortunately, this is usually not trivial, but some promising examples already exist. The AlphaGo approach that is based on searching by means of MCTS in a neural network representation of the treated problem has been transfered to the chemical retrosynthesis problem \cite{segler2018planning} that consists of finding a synthesis path for a specific chemical component as depicted in Fig.~\ref{fig:chem}.
As for the synthesis problem, in contrast to playing Go, the set of feasible moves (possible reactions) is not given but has to be learned from data, the approach bears some similarity to MuZero \cite{schrittwieser2019mastering}. The idea to learn a forward model from data has been termed \emph{world program} \cite{segler2019world}.

Similarly, the same distributed RL system that OpenAI used to
train a team of five agents for Dota 2   \cite{berner2019dota}, was used to train a robot hand to perform dexterous in-hand manipulation  \cite{andrychowicz2020learning}. 

We believe Game AI research will continue to drive innovations in the world of AI and hope this review article will serve as a useful guide for researchers entering this exciting research field.  
 
\section*{Acknowledgements}

We would like to thank Mads Lassen, Rasmus Berg Palm, Niels Justesen, Georgios Yannakakis, Marwin Segler, and Christian Igel for comments on earlier drafts of this manuscript. 

\printbibliography

@article{lucas2006evolutionary,
  title={Evolutionary computation and games},
  author={Lucas, Simon M and Kendall, Graham},
  journal={IEEE Computational Intelligence Magazine},
  volume={1},
  number={1},
  pages={10--18},
  year={2006},
  publisher={IEEE}
}

@article{ecoffet2019go,
  author    = {Adrien Ecoffet and
               Joost Huizinga and
               Joel Lehman and
               Kenneth O. Stanley and
               Jeff Clune},
  title     = {Go-Explore: a New Approach for Hard-Exploration Problems},
  year      = {2019},
  url       = {http://arxiv.org/abs/1901.10995},
  archivePrefix = {arXiv},
  eprint    = {1901.10995},
  bibsource = {dblp computer science bibliography, https://dblp.org}
}

@article{mnih2015human,
  title={Human-level control through deep reinforcement learning},
  author={Mnih, Volodymyr and Kavukcuoglu, Koray and Silver, David and Rusu, Andrei A and Veness, Joel and Bellemare, Marc G and Graves, Alex and Riedmiller, Martin and Fidjeland, Andreas K and Ostrovski, Georg and others},
  journal={Nature},
  volume={518},
  number={7540},
  pages={529},
  year={2015},
  publisher={Nature Publishing Group}
}

@article{hausknecht2014neuroevolution,
  title={A neuroevolution approach to general atari game playing},
  author={Hausknecht, Matthew and Lehman, Joel and Miikkulainen, Risto and Stone, Peter},
  journal={IEEE Transactions on Computational Intelligence and AI in Games},
  volume={6},
  number={4},
  pages={355--366},
  year={2014},
  publisher={IEEE}
}

@inproceedings{togelius2009super,
  title={Super mario evolution},
  author={Togelius, Julian and Karakovskiy, Sergey and Koutn{\'\i}k, Jan and Schmidhuber, Jurgen},
  booktitle={2009 ieee symposium on computational intelligence and games},
  pages={156--161},
  year={2009},
  organization={IEEE}
}

@article{summerville2018procedural,
  title={Procedural content generation via machine learning (PCGML)},
  author={Summerville, Adam and Snodgrass, Sam and Guzdial, Matthew and Holmg{\aa}rd, Christoffer and Hoover, Amy K and Isaksen, Aaron and Nealen, Andy and Togelius, Julian},
  journal={IEEE Transactions on Games},
  volume={10},
  number={3},
  pages={257--270},
  year={2018},
  publisher={IEEE}
}

@inproceedings{giacomello2018doom,
  title={DOOM level generation using generative adversarial networks},
  author={Giacomello, Edoardo and Lanzi, Pier Luca and Loiacono, Daniele},
  booktitle={2018 IEEE Games, Entertainment, Media Conference (GEM)},
  pages={316--323},
  year={2018},
  organization={IEEE}
}

@inproceedings{volz2018evolving,
  title={Evolving mario levels in the latent space of a deep convolutional generative adversarial network},
  author={Volz, Vanessa and Schrum, Jacob and Liu, Jialin and Lucas, Simon M and Smith, Adam and Risi, Sebastian},
  booktitle={Proceedings of the Genetic and Evolutionary Computation Conference},
  pages={221--228},
  year={2018},
  organization={ACM}
}

@inproceedings{gallagher2007evolving,
  title={Evolving pac-man players: Can we learn from raw input?},
  author={Gallagher, Marcus and Ledwich, Mark},
  booktitle={2007 IEEE Symposium on Computational Intelligence and Games},
  pages={282--287},
  year={2007},
  organization={IEEE}
}

@article{parker2012neurovisual,
  title={Neurovisual control in the Quake II environment},
  author={Parker, Matt and Bryant, Bobby D},
  journal={IEEE Transactions on Computational Intelligence and AI in Games},
  volume={4},
  number={1},
  pages={44--54},
  year={2012},
  publisher={IEEE}
}

@article{yannakakis2011experience,
  title={Experience-driven procedural content generation},
  author={Yannakakis, Georgios N and Togelius, Julian},
  journal={IEEE Transactions on Affective Computing},
  volume={2},
  number={3},
  pages={147--161},
  year={2011},
  publisher={IEEE}
}

@inproceedings{makantasis2019pixels,
  title={From Pixels to Affect: A Study on Games and Player Experience},
  author={Makantasis, Konstantinos and Liapis, Antonios and Yannakakis, Georgios N},
  booktitle={2019 8th International Conference on Affective Computing and Intelligent Interaction (ACII)},
  pages={1--7},
  year={2019},
  organization={IEEE}
}

@inproceedings{liapis2013sentient,
  title={Sentient Sketchbook: Computer-aided game level authoring.},
  author={Liapis, Antonios and Yannakakis, Georgios N and Togelius, Julian},
  booktitle={FDG},
  pages={213--220},
  year={2013}
}

@article{arulkumaran2017brief,
  title={A brief survey of deep reinforcement learning},
  author={Arulkumaran, Kai and Deisenroth, Marc Peter and Brundage, Miles and Bharath, Anil Anthony},
  journal={arXiv:1708.05866},
  year={2017}
}

@article{akkaya2019solving,
  title={Solving Rubik's Cube with a Robot Hand},
  author={Akkaya, Ilge and Andrychowicz, Marcin and Chociej, Maciek and Litwin, Mateusz and McGrew, Bob and Petron, Arthur and Paino, Alex and Plappert, Matthias and Powell, Glenn and Ribas, Raphael and others},
  journal={arXiv:1910.07113},
  year={2019}
}

@article{lecun2015deep,
  title={Deep learning},
  author={LeCun, Yann and Bengio, Yoshua and Hinton, Geoffrey},
  journal={nature},
  volume={521},
  number={7553},
  pages={436},
  year={2015},
  publisher={Nature Publishing Group}
}

@misc{alphastarblog,
  title="{AlphaStar: Mastering the Real-Time Strategy Game StarCraft II}",
  author={Vinyals, Oriol and Babuschkin, Igor and Chung, Junyoung and Mathieu, Michael and Jaderberg, Max and Czarnecki, Wojtek and Dudzik, Andrew and Huang, Aja and Georgiev, Petko and Powell, Richard and Ewalds, Timo and Horgan, Dan and Kroiss, Manuel and Danihelka, Ivo and Agapiou, John and Oh, Junhyuk and Dalibard, Valentin and Choi, David and Sifre, Laurent and Sulsky, Yury and Vezhnevets, Sasha and Molloy, James and Cai, Trevor and Budden, David and Paine, Tom and Gulcehre, Caglar and Wang, Ziyu and Pfaff, Tobias and Pohlen, Toby and Yogatama, Dani and Cohen, Julia and McKinney, Katrina and Smith, Oliver and Schaul, Tom and Lillicrap, Timothy and Apps, Chris and Kavukcuoglu, Koray and Hassabis, Demis and Silver, David},
  howpublished={\url{https://deepmind.com/blog/alphastar-mastering-real-time-strategy-game-starcraft-ii/}},
  year={2019}
}

@article{zhang2018study,
  title={A study on overfitting in deep reinforcement learning},
  author={Zhang, Chiyuan and Vinyals, Oriol and Munos, Remi and Bengio, Samy},
  journal={arXiv:1804.06893},
  year={2018}
}

@inproceedings{justesen2019we,
  title={When Are We Done with Games?},
  author={Justesen, Niels and Debus, Michael S and Risi, Sebastian},
  booktitle={2019 IEEE Conference on Games (CoG)},
  pages={1--8},
  year={2019},
  organization={IEEE}
}

@article{canaan2019leveling,
  title={Leveling the Playing Field-Fairness in AI Versus Human Game Benchmarks},
  author={Canaan, Rodrigo and Salge, Christoph and Togelius, Julian and Nealen, Andy},
  journal={arXiv:1903.-07008},
  year={2019}
}

@article{justesen2018illuminating,
  title={Illuminating generalization in deep reinforcement learning through procedural level generation},
  author={Justesen, Niels and Torrado, Ruben Rodriguez and Bontrager, Philip and Khalifa, Ahmed and Togelius, Julian and Risi, Sebastian},
  journal={arXiv:1806.10729},
  year={2018}
}

@ARTICLE{justesen19review,
author={N. {Justesen} and P. {Bontrager} and J. {Togelius} and S. {Risi}},
journal={IEEE Transactions on Games},
title={Deep Learning for Video Game Playing},
year={2019},
volume={},
number={},
pages={1-1},
doi={10.1109/TG.2019.2896986},
ISSN={2475-1510},
month={},}

@article{risi2015neuroevolution,
  title={Neuroevolution in games: State of the art and open challenges},
  author={Risi, Sebastian and Togelius, Julian},
  journal={IEEE Transactions on Computational Intelligence and AI in Games},
  volume={9},
  number={1},
  pages={25--41},
  year={2015},
  publisher={IEEE}
}

@misc{risi2019procedural,
    title={Procedural Content Generation: From Automatically Generating Game Levels to Increasing Generality in Machine Learning},
    author={Sebastian Risi and Julian Togelius},
    year={2019},
    eprint={1911.13071},
    archivePrefix={arXiv},
    primaryClass={cs.AI}
}

@article{YannakakisT11,
  author    = {Georgios N. Yannakakis and
               Julian Togelius},
  title     = {The 2010 {IEEE} Conference on Computational Intelligence and Games
               Report},
  journal   = {{IEEE} Comp. Int. Mag.},
  volume    = {6},
  number    = {2},
  pages     = {10--14},
  year      = {2011},
  doi       = {10.1109/MCI.2011.940612}
}

@article{YannakakisT15,
  author    = {Georgios N. Yannakakis and
               Julian Togelius},
  title     = {A Panorama of Artificial and Computational Intelligence in Games},
  journal   = {{IEEE} Trans. Comput. Intellig. and {AI} in Games},
  volume    = {7},
  number    = {4},
  pages     = {317--335},
  year      = {2015},
  url       = {https://doi.org/10.1109/TCIAIG.2014.2339221},
  doi       = {10.1109/TCIAIG.2014.2339221},
  bibsource = {dblp computer science bibliography, https://dblp.org}
}

@book{PCGbook,
  author    = {Noor Shaker and
               Julian Togelius and
               Mark J. Nelson},
  title     = {Procedural Content Generation in Games},
  series    = {Computational Synthesis and Creative Systems},
  publisher = {Springer},
  year      = {2016},
  url       = {https://doi.org/10.1007/978-3-319-42716-4},
  doi       = {10.1007/978-3-319-42716-4},
  isbn      = {978-3-319-42714-0},
  bibsource = {dblp computer science bibliography, https://dblp.org}
}

@article{Livingstone2006,
 author = {Livingstone, Daniel},
 title = {Turing's Test and Believable AI in Games},
 journal = {Comput. Entertain.},
 volume = {4},
 number = {1},
 year = {2006},
 issn = {1544-3574},
 doi = {10.1145/1111293.1111303},
 publisher = {ACM},
 address = {New York, NY, USA}
}

@article{such2017deep,
  title={Deep neuroevolution: Genetic algorithms are a competitive alternative for training deep neural networks for reinforcement learning},
  author={Such, Felipe Petroski and Madhavan, Vashisht and Conti, Edoardo and Lehman, Joel and Stanley, Kenneth O and Clune, Jeff},
  journal={arXiv: 1712.06567},
  year={2017}
}

@book{gameAIbook,
  author    = {Georgios N. Yannakakis and
               Julian Togelius},
  title     = {Artificial Intelligence and Games},
  publisher = {Springer},
  year      = {2018},
  url       = {https://doi.org/10.1007/978-3-319-63519-4},
  doi       = {10.1007/978-3-319-63519-4},
  isbn      = {978-3-319-63518-7},
  bibsource = {dblp computer science bibliography, https://dblp.org}
}

@article{Schaeffer2007,
	author = {Schaeffer, Jonathan and Burch, Neil and Bj{\"o}rnsson, Yngvi and Kishimoto, Akihiro and M{\"u}ller, Martin and Lake, Robert and Lu, Paul and Sutphen, Steve},
	title = {Checkers Is Solved},
	volume = {317},
	number = {5844},
	pages = {1518--1522},
	year = {2007},
	doi = {10.1126/science.1144079},
	publisher = {American Association for the Advancement of Science},
	issn = {0036-8075},
	URL = {https://science.sciencemag.org/content/317/5844/1518},
	eprint = {https://science.sciencemag.org/content/317/5844/1518.full.pdf},
	journal = {Science},
}

@inproceedings{Coulom06,
  author    = {R{\'{e}}mi Coulom},
  title     = {Efficient Selectivity and Backup Operators in Monte-Carlo Tree Search},
  booktitle = {Computers and Games, 5th International Conference, {CG} 2006, Turin, Italy, May 29-31, 2006. Revised Papers},
  pages     = {72--83},
  year      = {2006},
  editor    = {H. Jaap van den Herik and
               Paolo Ciancarini and
               H. H. L. M. Donkers},
  series    = {Lecture Notes in Computer Science},
  volume    = {4630},
  publisher = {Springer},
  doi       = {10.1007/978-3-540-75538-8},
  isbn      = {978-3-540-75537-1},
}

@inproceedings{denzinger2005dealing,
  title={Dealing with Parameterized Actions in Behavior Testing of Commercial Computer Games.},
  author={Denzinger, J{\"o}rg and Loose, Kevin and Gates, Darryl and Buchanan, John W},
  booktitle={CIG},
  year={2005},
  organization={Citeseer}
}

@article{ha2018world,
  title={World models},
  author={Ha, David and Schmidhuber, J{\"u}rgen},
  journal={arXiv: 1803.10122},
  year={2018}
}

@inproceedings{risi2019gecco,
 author = {Risi, Sebastian and Stanley, Kenneth O.},
 title = {Deep Neuroevolution of Recurrent and Discrete World Models},
 year = {2019},
 isbn = {9781450361118},
 publisher = {Association for Computing Machinery},
 address = {New York, NY, USA},
 url = {https://doi.org/10.1145/3321707.3321817},
 doi = {10.1145/3321707.3321817},
 booktitle = {Proceedings of the Genetic and Evolutionary Computation Conference},
 pages = {456–462},
 numpages = {7},
 location = {Prague, Czech Republic},
 series = {GECCO ’19}
}

@inproceedings{drachen2009player,
  title={Player modeling using self-organization in Tomb Raider: Underworld},
  author={Drachen, Anders and Canossa, Alessandro and Yannakakis, Georgios N},
  booktitle={2009 IEEE symposium on computational intelligence and games},
  pages={1--8},
  year={2009},
  organization={IEEE}
}

@article{andrychowicz2020learning,
  title={Learning dexterous in-hand manipulation},
  author={Andrychowicz, OpenAI: Marcin and Baker, Bowen and Chociej, Maciek and Jozefowicz, Rafal and McGrew, Bob and Pachocki, Jakub and Petron, Arthur and Plappert, Matthias and Powell, Glenn and Ray, Alex and others},
  journal={The International Journal of Robotics Research},
  volume={39},
  number={1},
  pages={3--20},
  year={2020},
  publisher={SAGE Publications Sage UK: London, England}
}

@article{berner2019dota,
  title={Dota 2 with Large Scale Deep Reinforcement Learning},
  author={Berner, Christopher and Brockman, Greg and Chan, Brooke and Cheung, Vicki and D{\k{e}}biak, Przemys{\l}aw and Dennison, Christy and Farhi, David and Fischer, Quirin and Hashme, Shariq and Hesse, Chris and others},
  journal={arXiv:1912.06680},
  year={2019}
}

@article{melhart2019your,
  title={Your Gameplay Says it All: Modelling Motivation in Tom Clancy's The Division},
  author={Melhart, David and Azadvar, Ahmad and Canossa, Alessandro and Liapis, Antonios and Yannakakis, Georgios N},
  journal={arXiv:1902.00040},
  year={2019}
}

@article{pedersen2010modeling,
  title={Modeling player experience for content creation},
  author={Pedersen, Christopher and Togelius, Julian and Yannakakis, Georgios N},
  journal={IEEE Transactions on Computational Intelligence and AI in Games},
  volume={2},
  number={1},
  pages={54--67},
  year={2010},
  publisher={IEEE}
}

@article{bakkes2012player,
  title={Player behavioural modelling for video games},
  author={Bakkes, Sander CJ and Spronck, Pieter HM and van Lankveld, Giel},
  journal={Entertainment Computing},
  volume={3},
  number={3},
  pages={71--79},
  year={2012},
  publisher={Elsevier}
}

@inproceedings{yannakakis2013player,
  title={Player modeling},
  author={Yannakakis, Georgios N and Spronck, Pieter and Loiacono, Daniele and Andr{\'e}, Elisabeth},
  year={2013},
  organization={Schloss Dagstuhl-Leibniz-Zentrum fuer Informatik}
}

@inproceedings{runge2014churn,
  title={Churn prediction for high-value players in casual social games},
  author={Runge, Julian and Gao, Peng and Garcin, Florent and Faltings, Boi},
  booktitle={2014 IEEE conference on Computational Intelligence and Games},
  pages={1--8},
  year={2014},
  organization={IEEE}
}

@article{risi2019improving,
  title={Improving Deep Neuroevolution via Deep Innovation Protection},
  author={Risi, Sebastian and Stanley, Kenneth O},
  journal={arXiv: 2001. 01683},
  year={2019}
}

@article{hafner2018learning,
  title={Learning latent dynamics for planning from pixels},
  author={Hafner, Danijar and Lillicrap, Timothy and Fischer, Ian and Villegas, Ruben and Ha, David and Lee, Honglak and Davidson, James},
  journal={arXiv:1811.04551},
  year={2018}
}

@article{stanley2019designing,
  title={Designing neural networks through neuroevolution},
  author={Stanley, Kenneth O and Clune, Jeff and Lehman, Joel and Miikkulainen, Risto},
  journal={Nature Machine Intelligence},
  volume={1},
  number={1},
  pages={24--35},
  year={2019},
  publisher={Nature Publishing Group}
}

@inproceedings{zhu2018explainable,
  title={Explainable AI for designers: A human-centered perspective on mixed-initiative co-creation},
  author={Zhu, Jichen and Liapis, Antonios and Risi, Sebastian and Bidarra, Rafael and Youngblood, G Michael},
  booktitle={2018 IEEE Conference on Computational Intelligence and Games (CIG)},
  pages={1--8},
  year={2018},
  organization={IEEE}
}

@inproceedings{hadiji2014predicting,
  title={Predicting player churn in the wild},
  author={Hadiji, Fabian and Sifa, Rafet and Drachen, Anders and Thurau, Christian and Kersting, Kristian and Bauckhage, Christian},
  booktitle={2014 IEEE Conference on Computational Intelligence and Games},
  pages={1--8},
  year={2014},
  organization={IEEE}
}

@book{el2016game,
  title={Game analytics},
  author={El-Nasr, Magy Seif and Drachen, Anders and Canossa, Alessandro},
  year={2016},
  publisher={Springer}
}

@inproceedings{perez2016general,
  title={General video game ai: Competition, challenges and opportunities},
  author={Perez-Liebana, Diego and Samothrakis, Spyridon and Togelius, Julian and Schaul, Tom and Lucas, Simon M},
  booktitle={Thirtieth AAAI Conference on Artificial Intelligence},
  year={2016}
}

@article{Browne2012,
author={C. B. {Browne} and E. {Powley} and D. {Whitehouse} and S. M. {Lucas} and P. I. {Cowling} and P. {Rohlfshagen} and S. {Tavener} and D. {Perez} and S. {Samothrakis} and S. {Colton}},
journal={IEEE Transactions on Computational Intelligence and AI in Games},
title={A Survey of Monte Carlo Tree Search Methods},
year={2012},
volume={4},
number={1},
pages={1-43},
doi={10.1109/TCIAIG.2012.2186810},
ISSN={1943-0698},
month={March},}

@article{Silver2016,
       author = {{Silver}, David and {Huang}, Aja and {Maddison}, Chris J. and
         {Guez}, Arthur and {Sifre}, Laurent and {van den Driessche}, George and
         {Schrittwieser}, Julian and {Antonoglou}, Ioannis and
         {Panneershelvam}, Veda and {Lanctot}, Marc and {Dieleman}, Sander and
         {Grewe}, Dominik and {Nham}, John and {Kalchbrenner}, Nal and
         {Sutskever}, Ilya and {Lillicrap}, Timothy and {Leach}, Madeleine and
         {Kavukcuoglu}, Koray and {Graepel}, Thore and {Hassabis}, Demis},
        title = "{Mastering the game of Go with deep neural networks and tree search}",
      journal = {Nature},
         year = "2016",
        month = "Jan",
       volume = {529},
       number = {7587},
        pages = {484-489},
          doi = {10.1038/nature16961},
}

@inproceedings{gudmundsson2018human,
  title={Human-like playtesting with deep learning},
  author={Gudmundsson, Stefan Freyr and Eisen, Philipp and Poromaa, Erik and Nodet, Alex and Purmonen, Sami and Kozakowski, Bartlomiej and Meurling, Richard and Cao, Lele},
  booktitle={2018 IEEE Conference on Computational Intelligence and Games (CIG)},
  pages={1--8},
  year={2018},
  organization={IEEE}
}

@book{Eiben2015,
author = {Eiben, A. E. and Smith, James E.},
title = {Introduction to Evolutionary Computing},
year = {2015},
isbn = {3662448734},
publisher = {Springer},
edition = {2nd}
}

@book{Plaat2020,
author = {Aske Plaat},
year = {2020},
title = {Learning to Play --- Reinforcement Learning and Games},
note = {https://learningtoplay.net/},
}

@article{silver2017mastering,
  author = {Silver, David and Schrittwieser, Julian and Simonyan, Karen and Antonoglou, Ioannis and Huang, Aja and Guez, Arthur and Hubert, Thomas and Baker, Lucas and Lai, Matthew and Bolton, Adrian and Chen, Yutian and Lillicrap, Timothy and Hui, Fan and Sifre, Laurent and van den Driessche, George and Graepel, Thore and Hassabis, Demis},
  journal = {Nature},
  month = oct,
  pages = {354-359},
  publisher = {Macmillan Publishers Limited, part of Springer Nature. All rights reserved.},
  title = {Mastering the game of Go without human knowledge},
  url = {http://dx.doi.org/10.1038/nature24270},
  volume = 550,
  year = 2017
}

@article{jaderberg2019human,
  title={Human-level performance in 3D multiplayer games with population-based reinforcement learning},
  author={Jaderberg, Max and Czarnecki, Wojciech M and Dunning, Iain and Marris, Luke and Lever, Guy and Castaneda, Antonio Garcia and Beattie, Charles and Rabinowitz, Neil C and Morcos, Ari S and Ruderman, Avraham and others},
  journal={Science},
  volume={364},
  number={6443},
  pages={859--865},
  year={2019},
  publisher={American Association for the Advancement of Science}
}

@article{Powley2014,
title = "Information capture and reuse strategies in Monte Carlo Tree Search, with applications to games of hidden information",
journal = "Artificial Intelligence",
volume = "217",
pages = "92 - 116",
year = "2014",
issn = "0004-3702",
doi = "https://doi.org/10.1016/j.artint.2014.08.002",
url = "http://www.sciencedirect.com/science/article/pii/S0004370214001052",
author = "Edward J. Powley and Peter I. Cowling and Daniel Whitehouse",
}

@article{Gelly2011,
title = "Monte-Carlo tree search and rapid action value estimation in computer Go",
journal = "Artificial Intelligence",
volume = "175",
number = "11",
pages = "1856 - 1875",
year = "2011",
issn = "0004-3702",
doi = "https://doi.org/10.1016/j.artint.2011.03.007",
author = "Sylvain Gelly and David Silver",
}

@article{Pepels2014,
  author    = {Tom Pepels and
               Mark H. M. Winands and
               Marc Lanctot},
  title     = {Real-Time Monte Carlo Tree Search in Ms Pac-Man},
  journal   = {{IEEE} Trans. Comput. Intellig. and {AI} in Games},
  volume    = {6},
  number    = {3},
  pages     = {245--257},
  year      = {2014},
  doi       = {10.1109/TCIAIG.2013.2291577},
}

@article{campbell2002deep,
  title={Deep blue},
  author={Campbell, Murray and Hoane Jr, A Joseph and Hsu, Feng-hsiung},
  journal={Artificial intelligence},
  volume={134},
  number={1-2},
  pages={57--83},
  year={2002},
  publisher={Elsevier}
}

@article{genesereth2005general,
  title={General game playing: Overview of the AAAI competition},
  author={Genesereth, Michael and Love, Nathaniel and Pell, Barney},
  journal={AI magazine},
  volume={26},
  number={2},
  pages={62--62},
  year={2005}
}

@article {Silver2018,
	author = {Silver, David and Hubert, Thomas and Schrittwieser, Julian and Antonoglou, Ioannis and Lai, Matthew and Guez, Arthur and Lanctot, Marc and Sifre, Laurent and Kumaran, Dharshan and Graepel, Thore and Lillicrap, Timothy and Simonyan, Karen and Hassabis, Demis},
	title = {A general reinforcement learning algorithm that masters chess, shogi, and Go through self-play},
	volume = {362},
	number = {6419},
	pages = {1140--1144},
	year = {2018},
	doi = {10.1126/science.aar6404},
	publisher = {American Association for the Advancement of Science},
	issn = {0036-8075},
	URL = {https://science.sciencemag.org/content/362/6419/1140},
    journal = {Science}
}

@misc{schrittwieser2019mastering,
    title={Mastering Atari, Go, Chess and Shogi by Planning with a Learned Model},
    author={Julian Schrittwieser and Ioannis Antonoglou and Thomas Hubert and Karen Simonyan and Laurent Sifre and Simon Schmitt and Arthur Guez and Edward Lockhart and Demis Hassabis and Thore Graepel and Timothy Lillicrap and David Silver},
    year={2019},
    eprint={1911.08265},
    archivePrefix={arXiv},
    primaryClass={cs.LG}
}

@article {Brown2019,
	author = {Brown, Noam and Sandholm, Tuomas},
	title = {Superhuman AI for multiplayer poker},
	volume = {365},
	number = {6456},
	pages = {885--890},
	year = {2019},
	doi = {10.1126/science.aay2400},
	publisher = {American Association for the Advancement of Science},
	issn = {0036-8075},
	URL = {https://science.sciencemag.org/content/365/6456/885},
	journal = {Science}
}

@misc{Lerer2019,
    title={Improving Policies via Search in Cooperative Partially Observable Games},
    author={Adam Lerer and Hengyuan Hu and Jakob Foerster and Noam Brown},
    year={2019},
    eprint={1912.02318},
    archivePrefix={arXiv},
    primaryClass={cs.AI}
}

@article{vinyals2019,
author = {Vinyals, Oriol and Babuschkin, Igor and Czarnecki, Wojciech and Mathieu, Michaël and Dudzik, Andrew and Chung, Junyoung and Choi, David and Powell, Richard and Ewalds, Timo and Georgiev, Petko and Oh, Junhyuk and Horgan, Dan and Kroiss, Manuel and Danihelka, Ivo and Huang, Aja and Sifre, Laurent and Cai, Trevor and Agapiou, John and Jaderberg, Max and Silver, David},
year = {2019},
month = {11},
pages = {},
title = {Grandmaster level in StarCraft II using multi-agent reinforcement learning},
volume = {575},
journal = {Nature},
doi = {10.1038/s41586-019-1724-z}
}

@misc{Baker2019,
    title={Emergent Tool Use From Multi-Agent Autocurricula},
    author={Bowen Baker and Ingmar Kanitscheider and Todor Markov and Yi Wu and Glenn Powell and Bob McGrew and Igor Mordatch},
    year={2019},
    eprint={1909.07528},
    archivePrefix={arXiv},
}

@inproceedings{Juliani2019,
  author    = {Arthur Juliani and
               Ahmed Khalifa and
               Vincent{-}Pierre Berges and
               Jonathan Harper and
               Ervin Teng and
               Hunter Henry and
               Adam Crespi and
               Julian Togelius and
               Danny Lange},
  title     = {Obstacle Tower: {A} Generalization Challenge in Vision, Control, and Planning},
  booktitle = {Proceedings of the Twenty-Eighth International Joint Conference on
               Artificial Intelligence, {IJCAI} 2019, Macao, China, August 10-16,
               2019},
  pages     = {2684--2691},
  year      = {2019},
  doi       = {10.24963/ijcai.2019/373},
  editor    = {Sarit Kraus},
  publisher = {ijcai.org},
}

@article{segler2018planning,
  title={Planning chemical syntheses with deep neural networks and symbolic AI},
  author={Segler, Marwin HS and Preuss, Mike and Waller, Mark P},
  journal={Nature},
  volume={555},
  number={7698},
  pages={604},
  year={2018},
  publisher={Nature Publishing Group}
}

@inproceedings{goodfellow2014generative,
  title={Generative adversarial nets},
  author={Goodfellow, Ian and Pouget-Abadie, Jean and Mirza, Mehdi and Xu, Bing and Warde-Farley, David and Ozair, Sherjil and Courville, Aaron and Bengio, Yoshua},
  booktitle={Advances in neural information processing systems},
  pages={2672--2680},
  year={2014}
}

@article{Liapis2019,
  author    = {Antonios Liapis and
               Georgios N. Yannakakis and
               Mark J. Nelson and
               Mike Preuss and
               Rafael Bidarra},
  title     = {Orchestrating Game Generation},
  journal   = {{IEEE} Trans. Games},
  volume    = {11},
  number    = {1},
  pages     = {48--68},
  year      = {2019},
  url       = {https://doi.org/10.1109/TG.2018.2870876},
  doi       = {10.1109/TG.2018.2870876},
}

@article{Ontanon2013,
  author    = {Santiago Onta{\~{n}}{\'{o}}n and
               Gabriel Synnaeve and
               Alberto Uriarte and
               Florian Richoux and
               David Churchill and
               Mike Preuss},
  title     = {A Survey of Real-Time Strategy Game {AI} Research and Competition
               in StarCraft},
  journal   = {{IEEE} Trans. Comput. Intellig. and {AI} in Games},
  volume    = {5},
  number    = {4},
  pages     = {293--311},
  year      = {2013},
  url       = {https://doi.org/10.1109/TCIAIG.2013.2286295},
  doi       = {10.1109/TCIAIG.2013.2286295},
}

@inproceedings{PerezSLR13,
  author    = {Diego Perez Liebana and
               Spyridon Samothrakis and
               Simon M. Lucas and
               Philipp Rohlfshagen},
  title     = {Rolling horizon evolution versus tree search for navigation in single-player
               real-time games},
  booktitle = {Genetic and Evolutionary Computation Conference, {GECCO} '13, Amsterdam,
               The Netherlands, July 6-10, 2013},
  pages     = {351--358},
  year      = {2013},
  url       = {https://doi.org/10.1145/2463372.2463413},
  doi       = {10.1145/2463372.2463413},
  editor    = {Christian Blum and
               Enrique Alba},
  publisher = {{ACM}},
}

@inproceedings{kempka2016vizdoom,
  title={Vizdoom: A doom-based ai research platform for visual reinforcement learning},
  author={Kempka, Micha{\l} and Wydmuch, Marek and Runc, Grzegorz and Toczek, Jakub and Ja{\'s}kowski, Wojciech},
  booktitle={2016 IEEE Conference on Computational Intelligence and Games (CIG)},
  pages={1--8},
  year={2016},
  organization={IEEE}
}

@inproceedings{liu2019,
title={Emergent Coordination Through Competition},
author={Siqi Liu and Guy Lever and Nicholas Heess and Josh Merel and Saran Tunyasuvunakool and Thore Graepel},
booktitle={International Conference on Learning Representations},
year={2019},
}

@inproceedings{PreussPVP18,
  author    = {Mike Preuss and
               Thomas Pfeiffer and
               Vanessa Volz and
               Nicolas Pflanzl},
  title     = {Integrated Balancing of an {RTS} Game: Case Study and Toolbox Refinement},
  booktitle = {2018 {IEEE} Conference on Computational Intelligence and Games, {CIG}
               2018, Maastricht, The Netherlands, August 14-17, 2018},
  pages     = {1--8},
  year      = {2018},
  publisher = {{IEEE}},
}

@inproceedings{Kummer2018,
  author    = {Luiz Bernardo Martins Kummer and
               J{\'{u}}lio C{\'{e}}sar Nievola and
               Emerson Cabrera Paraiso},
  title     = {Applying Commitment to Churn and Remaining Players Lifetime Prediction},
  booktitle = {2018 {IEEE} Conference on Computational Intelligence and Games, {CIG}
               2018, Maastricht, The Netherlands, August 14-17, 2018},
  pages     = {1--8},
  publisher = {{IEEE}},
  year      = {2018},
}

@Article{Volz2019,
author="Volz, Vanessa",
title="Uncertainty Handling in Surrogate Assisted Optimisation of Games",
journal="KI - K{\"u}nstliche Intelligenz",
year="2019",
url="https://doi.org/10.1007/s13218-019-00613-1"
}

@article{spronck2006adaptive,
  title={Adaptive game AI with dynamic scripting},
  author={Spronck, Pieter and Ponsen, Marc and Sprinkhuizen-Kuyper, Ida and Postma, Eric},
  journal={Machine Learning},
  volume={63},
  number={3},
  pages={217--248},
  year={2006},
  publisher={Springer}
}

@book{hingston2012,
author = {Hingston, Philip},
title = {Believable Bots: Can Computers Play Like People?},
year = {2012},
isbn = {3642323227},
publisher = {Springer}
}

@inproceedings{torrado2018,
title = "Deep Reinforcement Learning for General Video Game AI",
author = "Torrado, {Ruben Rodriguez} and Philip Bontrager and Julian Togelius and Jialin Liu and Diego Perez-Liebana",
year = "2018",
month = "10",
day = "11",
doi = "10.1109/CIG.2018.8490422",
booktitle = "Proceedings of the 2018 IEEE Conference on Computational Intelligence and Games, CIG 2018",
publisher = "IEEE",
}

@article{yannakakis2014emotion,
  title={Emotion in games},
  author={Yannakakis, Georgios N and Paiva, Ana},
  journal={Handbook on affective computing},
  pages={459--471},
  year={2014},
  publisher={Oxford University Press}
}

@article{yannakakis2018ordinal,
  title={The ordinal nature of emotions: An emerging approach},
  author={Yannakakis, Georgios N and Cowie, Roddy and Busso, Carlos},
  journal={IEEE Transactions on Affective Computing},
  year={2018},
  publisher={IEEE}
}

@article{newellturing,
  address = {New York, NY, USA},
  author = {Newell, Allen and Simon, Herbert A.},
  book = {ACM Turing award lectures},
  doi = {10.1145/1283920.1283930},
  isbn = {0-201-0779X-X},
  journal = {Communications of the ACM},
  month = {March},
  number = 3,
  pages = {113-126},
  publisher = {ACM},
  title = {Computer science as empirical inquiry: symbols and search},
  url = {http://doi.acm.org/10.1145/1283920.1283930},
  volume = 19,
  year = 1976
}

@article{Bard2019,
  author    = {Nolan Bard and
               Jakob N. Foerster and
               Sarath Chandar and
               Neil Burch and
               Marc Lanctot and
               H. Francis Song and
               Emilio Parisotto and
               Vincent Dumoulin and
               Subhodeep Moitra and
               Edward Hughes and
               Iain Dunning and
               Shibl Mourad and
               Hugo Larochelle and
               Marc G. Bellemare and
               Michael Bowling},
  title     = {The Hanabi Challenge: {A} New Frontier for {AI} Research},
  journal   = {CoRR},
  volume    = {abs/1902.00506},
  year      = {2019},
  url       = {http://arxiv.org/abs/1902.00506},
  archivePrefix = {arXiv},
  eprint    = {1902.00506},
  bibsource = {dblp computer science bibliography, https://dblp.org}
}

@article{Machado2017,
  author    = {Marlos C. Machado and
               Marc G. Bellemare and
               Erik Talvitie and
               Joel Veness and
               Matthew J. Hausknecht and
               Michael Bowling},
  title     = {Revisiting the Arcade Learning Environment: Evaluation Protocols and
               Open Problems for General Agents},
  journal   = {CoRR},
  volume    = {abs/1709.06009},
  year      = {2017},
  url       = {http://arxiv.org/abs/1709.06009},
  archivePrefix = {arXiv},
  eprint    = {1709.06009},
  bibsource = {dblp computer science bibliography, https://dblp.org}
}

@article{Mnih2013,
  author    = {Volodymyr Mnih and
               Koray Kavukcuoglu and
               David Silver and
               Alex Graves and
               Ioannis Antonoglou and
               Daan Wierstra and
               Martin A. Riedmiller},
  title     = {Playing Atari with Deep Reinforcement Learning},
  journal   = {CoRR},
  volume    = {abs/1312.5602},
  year      = {2013},
  url       = {http://arxiv.org/abs/1312.5602},
  archivePrefix = {arXiv},
}

@misc{segler2019world,
    title={World Programs for Model-Based Learning and Planning in Compositional State and Action Spaces},
    author={Marwin H. S. Segler},
    year={2019},
    eprint={1912.13007},
    archivePrefix={arXiv},
    primaryClass={cs.LG}
}

@book{bartz2010experimental,
  title={Experimental methods for the analysis of optimization algorithms},
  author={Bartz-Beielstein, Thomas and Chiarandini, Marco and Paquete, Lu{\'\i}s and Preuss, Mike},
  year={2010},
  publisher={Springer}
}

@misc{ OpenAI_dota,
      author = {OpenAI},
      title = {OpenAI Five},
      howpublished = {\url{https://blog.openai.com/openai-five/}},
      year = {2018}
}

@article{raiman2019long,
  title={Long-Term Planning and Situational Awareness in OpenAI Five},
  author={Raiman, Jonathan and Zhang, Susan and Wolski, Filip},
  journal={arXiv preprint arXiv:1912.06721},
  year={2019}
}

@ARTICLE{Nareyek2007,
author={A. {Nareyek}},
journal={IEEE Intelligent Systems},
title={Game AI is Dead. Long Live Game AI!},
year={2007},
volume={22},
number={1},
pages={9-11},
doi={10.1109/MIS.2007.10},
ISSN={1941-1294},
month={Jan},}

@inproceedings{Yannakakis2012,
author = {Yannakakis, Geogios N.},
title = {Game AI Revisited},
year = {2012},
isbn = {9781450312158},
publisher = {Association for Computing Machinery},
address = {New York, NY, USA},
url = {https://doi.org/10.1145/2212908.2212954},
doi = {10.1145/2212908.2212954},
booktitle = {Proceedings of the 9th Conference on Computing Frontiers},
pages = {285–292},
numpages = {8},
location = {Cagliari, Italy},
series = {CF ’12}
}

@inproceedings{Mateas2003,
author={Michael Mateas},
year={2003},
title={Expressive AI: Games and Artificial Intelligence},
isbn={ISSN 2342-9666},
booktitle={DiGRA \&\#3903 - Proceedings of the 2003 DiGRA International Conference: Level Up},
url={http://www.digra.org/wp-content/uploads/digital-library/05150.02104.pdf}
}

@inproceedings{Buro2003,
  author    = {Michael Buro},
  title     = {Real-Time Strategy Games: {A} New {AI} Research Challenge},
  booktitle = {IJCAI-03, Proceedings of the Eighteenth International Joint Conference
               on Artificial Intelligence, Acapulco, Mexico, August 9-15, 2003},
  pages     = {1534--1535},
  year      = {2003},
  url       = {http://ijcai.org/Proceedings/03/Papers/265.pdf},
  editor    = {Georg Gottlob and
               Toby Walsh},
  publisher = {Morgan Kaufmann},
}

@InProceedings{Nareyek2001,
author="Nareyek, Alexander",
editor="Marsland, Tony
and Frank, Ian",
title="Review: Intelligent Agents for Computer Games",
booktitle="Computers and Games",
year="2001",
publisher="Springer Berlin Heidelberg",
address="Berlin, Heidelberg",
pages="414--422",
isbn="978-3-540-45579-0"
}

\end{document}